\documentclass[final]{cvpr}
\usepackage[pagebackref=true,breaklinks=true,
letterpaper=true,colorlinks,citecolor=blue,
linkcolor=blue,bookmarks=false]{hyperref}
\usepackage{cite}

\usepackage{times}
\usepackage{epsfig}
\usepackage{graphicx}
\usepackage{amsmath}
\usepackage{amssymb}
\usepackage{rotating}
\DeclareMathOperator*{\argmax}{argmax}
\DeclareMathOperator*{\argmin}{argmin}
\usepackage{multirow}
\newcommand{\edited}[1]{\textcolor{black}{#1}}
\newcommand{\Ja}{J_{2D}} 
\newcommand{\Jb}{J_{3D}} 
\newcommand{\Va}{V_{2D}} 
\newcommand{\Vb}{V_{3D}} 
\newcommand{\bJa}{\bar{J}_{2D}} 
\newcommand{\bJb}{\bar{J}_{3D}} 
\newcommand{\bVb}{\bar{V}_{3D}} 

\usepackage[pagebackref=true,breaklinks=true,colorlinks,bookmarks=false]{hyperref}

\def\cvprPaperID{6726} 
\def\confYear{CVPR 2022}

\begin{document}

\title{A Probabilistic Attention Model with Occlusion-aware Texture Regression for 3D Hand Reconstruction from a Single RGB Image}

\author{Zheheng Jiang\textsuperscript{1}
~~~ Hossein Rahmani\textsuperscript{1}
~~~ Sue Black\textsuperscript{2} 
~~~ Bryan M. Williams\textsuperscript{1}\\
\textsuperscript{1}Lancaster University ~~
\textsuperscript{2}St John's College of the University of Oxford ~~ \\
{\tt\small \{z.jiang11,h.rahmani,b.williams6\}@lancaster.ac.uk, sue.black@sjc.ox.ac.uk}\\ 
}

\maketitle

\begin{abstract}
\edited{
Recently, deep learning based approaches have shown promising results in 3D hand reconstruction from a single RGB image. These approaches can be roughly divided into 
model-based approaches, which are heavily dependent on the model's parameter space, and model-free approaches, 
which require large numbers of 3D ground truths to reduce depth ambiguity and struggle in weakly-supervised scenarios.
To overcome these issues, we propose a novel probabilistic model to achieve the robustness of model-based approaches and reduced dependence on the model's parameter space of model-free approaches. The proposed probabilistic model incorporates a model-based network as a prior-net to estimate the prior probability distribution of joints and vertices. 
An Attention-based Mesh Vertices Uncertainty Regression (AMVUR) model is proposed to capture dependencies among vertices and the correlation between joints and mesh vertices to improve their feature representation.
We further propose a learning based occlusion-aware Hand Texture Regression model to achieve high-fidelity texture reconstruction. 
We demonstrate the flexibility of the proposed probabilistic model to be trained in both supervised and weakly-supervised scenarios.
The experimental results demonstrate our probabilistic model's state-of-the-art accuracy in 3D hand and texture reconstruction from a single image in both training schemes, including in the presence of severe occlusions.}
\end{abstract}
%
\section{Introduction}
3D hand shape and texture reconstruction from a single RGB image is a challenging problem that has numerous applications such as human-machine interaction \cite{conci2007natural,yin2021wearable}, virtual and augmented reality \cite{jung2016body, wang2020rgb2hands,han2020megatrack,mueller2019real}, and sign language translation \cite{liang2020multi}. In recent years, there has been significant progress in reconstructing 3D hand pose and shape from a monocular images \cite{kolotouros2019convolutional,ge20193d,kulon2020weakly,choi2020pose2mesh,lin2021end,chen2021model,park2022handoccnet,boukhayma20193d,hampali2022keypoint}. These approaches can be generally categorized into model-based and model-free approaches. Model-based approaches \cite{ge20193d,chen2021model,park2022handoccnet,boukhayma20193d} utilize a parametric model such as MANO \cite{romero2017embodied} and train a network to regress its parametric representation in terms of shape and pose. Since the parametric model contains priors of human hands, these approaches are robust to environment variations and weakly-supervised training\cite{lin2021end}. However, the shape and pose regression is constrained by the parametric model that is learned from the limited hand exemplars \cite{kolotouros2019convolutional}. 

In contrast, model-free approaches \cite{lin2021end,kolotouros2019convolutional,choi2020pose2mesh, hampali2022keypoint} regress the coordinates of 3D hand joints and mesh directly instead of using parametric models. Despite the remarkable results they have achieved, there are several limitations. For example, Graph-CNN is used by \cite{choi2020pose2mesh,kolotouros2019convolutional} to model neighborhood vertex-vertex interactions, but such models cannot capture long range dependencies among vertices.
Although \cite{lin2021end} has addressed this issue by employing self-attention mechanism, it does not distinguish joints and vertices, processing them together in a same self-attention module.
Moreover none of these works can support weakly supervised training and often require a large amount of 3D annotations of both joints and vertices to reduce depth ambiguity in monocular 3D reconstruction \cite{spurr2020weakly}.

Motivated by the above observations, our first goal is to combine the benefits of the model-based and model-free approaches. To this end, we develop a probabilistic method that incorporates the MANO model into a prior-net to estimate the prior probability distribution of joints and vertices instead of using deterministic settings as previous approaches have done. To relax the solution space of the MANO model, an Attention-based Mesh Vertices Uncertainty Regression model (AMVUR) is proposed to estimate the conditioned probability distribution of the joints and vertices. In AMVUR, to improve feature representation of joints and vertices, a cross-attention model is proposed to capture the correlation between 3D positional encoded joints and mesh vertices, followed by a self-attention model for capturing the short/long range dependencies among mesh vertices. 
With the proposed architecture, the AMVUR model can be jointly trained with the prior-net to achieve superior performance to using them independently. To the best of our knowledge, our probabilistic attention model is the first approach that learns the probability distribution of hand joints and mesh under a probabilistic model.

The ability to reconstruct 3D hands with high-fidelity texture is helpful for 3D Hand Personalization and improves the performance of hand tracking systems \cite{de2008model,tkach2017online, taylor2017articulated}. Moreover, Hand texture reconstruction is important for the user experience and bodily self-consciousness in immersive virtual reality systems \cite{jung2016body}. We thus propose a learning based occlusion-aware hand texture regression model by introducing an occlusion-aware rasterization and reverse interpolation to achieve high-fidelity hand texture reconstruction. 

Our contributions are summarized as follows: 
\textbf{(1)} We introduce an Attention-based Mesh Vertices Uncertainty Regression model (AMVUR) comprising a cross attention module for capturing the correlation between joints and mesh vertices and a self-attention module for capturing the short/long range dependencies among mesh vertices.
\textbf{(2)} We propose a novel probabilistic attention model to learn the probability distribution of hand joints and mesh vertices, where the MANO parametric model is regarded as a prior-net and jointly trained with AMVUR.
\textbf{(3)} We propose an Occlusion-aware Hand Texture Regression model to achieve high-fidelity hand texture reconstruction, including in the presence of severe occlusions.
\textbf{(4)} We demonstrate that our network can be trained in both fully supervised and weakly supervised training schemes, achieving state-of-the-art (SOTA) performance on the three benchmark 3D hand reconstruction datasets: HO3Dv2 \cite{hampali2020honnotate}, HO3Dv3 \cite{hampali2021ho} and FreiHand \cite{zimmermann2019freihand}.

\begin{figure*}
\begin{center}
\includegraphics[width=13cm]{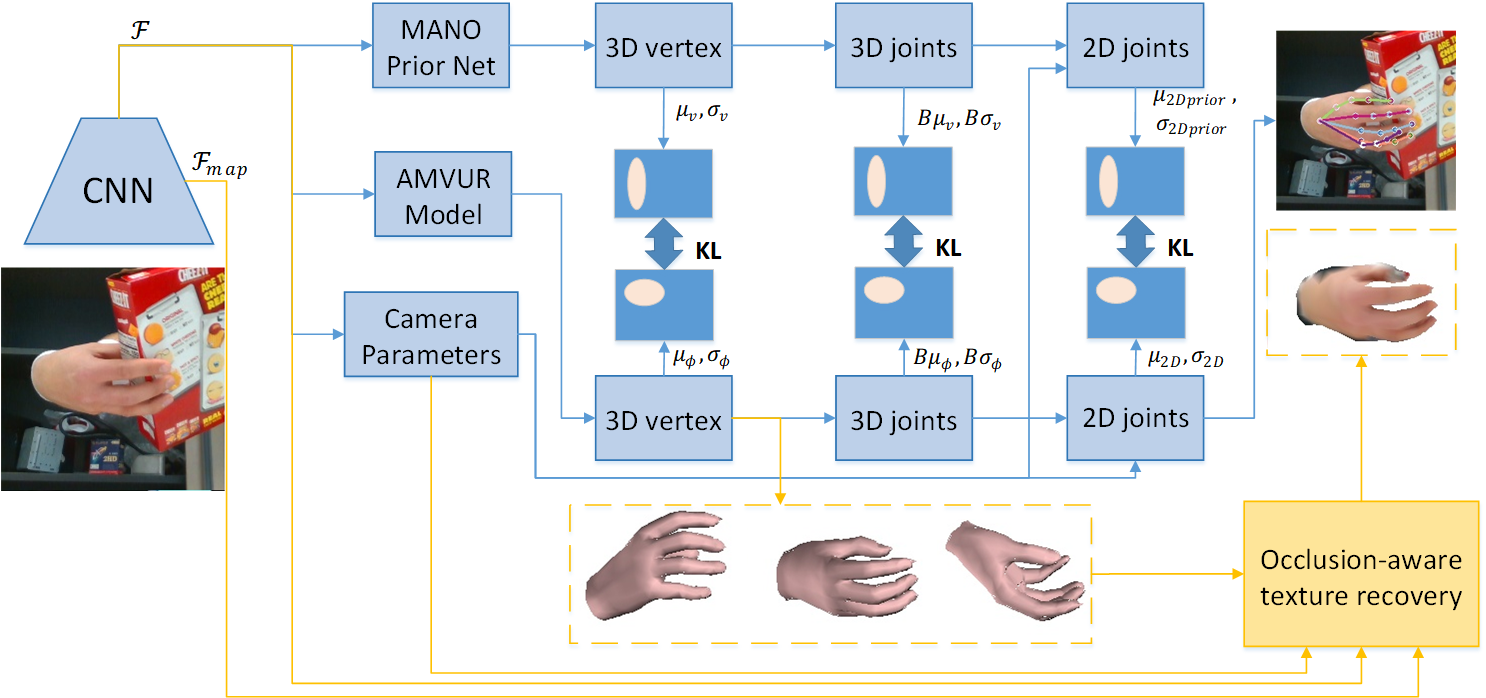}
\end{center}
\vspace{-1.0em}
\caption{Overview of our proposed method. The blue arrows and yellow arrows denote the flows of hand mesh reconstruction and hand texture regression respectively. We firstly extract a global feature vector $\mathcal{F}$ and a shallow feature map $\mathcal{F}_{map}$ from the backbone CNN. For hand mesh reconstruction, our MANO prior-net and 
AMVUR model take the global feature vector as input and are jointly trained to estimate the probability distributions of the 3D vertices, 3D joints and 2D joints. During training, the probability distributions estimated by these two models are tied by the KL-divergence. The camera model estimates camera parameters that are used to project the 3D joints and 3D vertices to 2D space. For hand texture regression, an Occlusion-aware texture recovery model is proposed to reconstruct occlusion-aware high-fidelity hand texture by taking the global feature vector, the shallow feature map, estimated camera parameters and estimated 3D vertices as inputs. During inference, the test image is passed through the CNN followed by AMVUR to generate the most likely 3D hand mesh which is then fed to the Occlusion-aware texture recovery model to reconstruct a textured mesh.
} 
\vspace{-1.0em}
\label{fig:framework}
\end{figure*}

\section{Related Work}
\noindent\textbf{Model-based Methods:} Recently, numerous works have been proposed to reconstruct the 3D hand by regressing the shape and pose parameters of a parametric hand model named MANO \cite{romero2017embodied} that is learned from around 1K high-resolution 3D hand scans. Boukhayma et al.\cite{boukhayma20193d} regress these parameters along with camera parameters via a deep convolutional encoder which takes a hand image and 2D joint heat-maps extracted from a joint detection network as input. Zhang et al. \cite{zhang2019end} propose an iterative regression module to fit the camera and model parameters from 2D joint heat-maps. Attention based approaches have also received increasing attention. Liu et al. \cite{liu2021semi} introduce an attention based contextual reasoning module for modeling hand-object interaction. A most recent approach \cite{park2022handoccnet} proposes to inject hand information into occluded regions by using a self-attention mechanism. However, their approach is a 2D spatial attention mechanism that is unable to capture correlation between mesh vertices in 3D space.

By utilizing strong hand priors of MANO, several other approaches \cite{spurr2020weakly,chen2021model} have attempted to reconstruct 3D hand shape and pose with weak supervision. Kulon et al. \cite{kulon2020weakly} apply Parametric Model Fitting to generate 3D mesh from detected 2D hand keypoints. The fitted mesh is then used as a supervisory signal to train a feed-forward network with a mesh convolutional decoder. Spurr et al. \cite{spurr2020weakly} introduce biomechanical constraints to guide the network to predict feasible hand poses with weakly-annotated real-world data. Chen et al. \cite{chen2021model} use 2D joints extracted from an off-the-shelf 2D pose estimator as a supervisory signal to train a model-based autoencoder to estimate 3D hand pose and shape.  However, similar to the model based approaches, they do not exploit correlation between joints and mesh vertices, yet our proposed AMVUR model addresses this issue and improves the feature representation of joints and vertices.

\noindent\textbf{Model-free Methods:} Although hand parametric models such as MANO serve as a strong structural prior to support 3D hand reconstruction, help to handle severe occlusions and help to accommodate weakly-annotated data, approaches that rely on this can easily get stuck in the model’s parameter space, resulting in a non-minimal representation problem \cite{kolotouros2019convolutional,choi2020pose2mesh}. To relax this heavy reliance on the parameter space, some approaches directly regress 3D positions of mesh vertices instead of predicting the model’s parameters. Among these approaches, Kolotouros et al. \cite{kolotouros2019convolutional} and Hongsuk et al. \cite{choi2020pose2mesh} combine an image-based CNN and a GraphCNN to estimate human mesh coordinates directly. Lin et al. \cite{lin2021end} argue that GraphCNN can only capture the local interactions between neighboring vertices of the triangle mesh, so they use a self-attention mechanism to capture global interactions between the vertices and joints. 
Most recently, Hampali et al. \cite{hampali2022keypoint} first extract joint features by localizing them on CNN feature maps, then take these features and their spatial encodings as the input to a transformer model for 3D hand pose estimation. However, spatial encoding is ambiguous to describe joints' 3D locations, especially for overlapping 3D joints in 2D images. 
Different from the above approaches, in AMVUR, a cross-attention module is proposed to learn the correlation between joints and mesh vertices, followed by a self-attention module to learn the correlation between different vertices. 

\noindent\textbf{Texture Reconstruction:} 3D hand texture estimation has wide applications in virtual and augmented reality, but most of the previously mentioned approaches do not address this problem. Qian et al. \cite{qian2020html} follow the 3D morphable face models \cite{dai20173d} to create a parametric hand texture model using principal component analysis (PCA). However this model requires a hand dataset with 3D textured scans and is established on only 51 hand exemplars. Recently, Chen et al. \cite{chen2021model} propose to regress hand texture via MLP layers by taking a global feature vector as input. However, this is limited to generating very coarse hand texture and is unable to recognize occlusions. In contrast, our texture hand model is able to reconstruct high-fidelity hand texture with occlusions.


\section{Proposed Model}
An overview of our proposed model is presented in Fig. \ref{fig:framework}. We first extract a global feature vector $\mathcal{F}$ and a shallow feature map $\mathcal{F}_{map}$ from our backbone Convolutional Neural Network (CNN). Then, we introduce a Bayesian model to describe the relationship between 3D hand joints, 3D hand mesh vertices, 2D hand joints and camera parameters for the regression task by taking $\mathcal{F}$ as input. This is described in Sections \ref{sec:problem_formulation} and \ref{sec:weakly-supervised} along with our problem formulation. Our proposed AMVUR uses $\mathcal{F}$ to learn the correlation between joints and mesh vertices, described in Section \ref{sec:AMVUR}. Our proposed Occlusion-aware Hand Texture Regression, which addresses occlusion-aware high-fidelity hand texture reconstruction by utilizing the global feature vector and the shallow feature map, is described in Section \ref{sec:OHTR}.

\subsection{Problem Formulation}\label{sec:problem_formulation}
\begin{figure}
\begin{center}
\includegraphics[width=4cm]{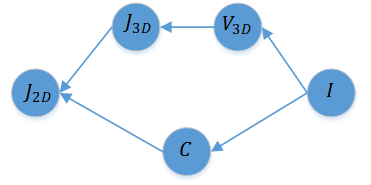}
\end{center}
\vspace{-1.2em}
\caption{A DAG describes the dependence between 2D joints, 3D joints, 3D mesh vertices, camera parameters and image.}
\vspace{-1.2em}
\label{fig:dependence}
\end{figure}

Given a 2D image $I$ containing a hand, our goal is to predict the locations of the 2D hand joints $\Ja\in \mathbb{R}^{K \times 2}$, 3D hand joints $\Jb\in \mathbb{R}^{K \times 3}$, 3D hand mesh vertices $\Vb\in \mathbb{R}^{\mathcal{V} \times 3}$ and camera parameters $C$, where $K$ is the number of joints and $\mathcal{V}$ is the number of mesh vertices. Most MANO based methods \cite{ge20193d,kulon2020weakly,chen2021model,park2022handoccnet} firstly propose a deep regression model to fit the MANO pose and shape parameters, from which the final 3D hand mesh is estimated via a MANO layer. This limits the learning ability of the deep neural network for 3D hand mesh regression. In contrast to the above methods, we use the MANO model as a prior-net and integrate it with our proposed Attention-based Mesh Vertices Uncertainty Regression model (AMVUR) for end-to-end training. Let $\delta$ and $\theta$ donate the model parameters of prior-net and AMVUR respectively, learned from the training dataset $D = \left\{\Jb^{i},\Vb^{i},\Ja^{i},C^{i},I^{i}\right\}^{T}_{i=1}$, where $T$ is the total number of training images. The model parameters are estimated by maximizing the log likelihood function
\begin{equation}
\small
    \ln\mathcal{L}\left( \delta \right)=\ln\prod_{i}P\left(\Ja^{i},\Jb^{i},\Vb^{i},C^{i}|I^{i};\delta\right)
\end{equation}
where $P\left(\Ja^{i},\Jb^{i},\Vb^{i},C^{i}|I^{i};\delta\right)$ is a prior joint probability distribution estimated using the MANO model. This is maximised as:
\begin{equation}
\small
\begin{split}
&\argmax_{\delta,\theta}\ln\mathcal{L}\left( \delta,\theta \right)\\
&=\argmin_{\delta,\theta}\sum_{i}\left(-\ln Q\left(\Ja^{i},\Jb^{i},\Vb^{i},C^{i}|I^{i};\theta\right) \right.\\
& \left. \quad +\ln \frac{Q\left(\Ja^{i},\Jb^{i},\Vb^{i},C^{i}|I^{i};\theta\right)}{P\left(\Ja^{i},\Jb^{i},\Vb^{i},C^{i}|I^{i};\delta\right)}\right),
\end{split}
\label{eq:MLE}
\end{equation}
where $Q\left(\Ja^{i},\Jb^{i},\Vb^{i},C^{i}|I^{i};\theta\right)$ is an approximate joint probability distribution that is learned by the proposed Attention-based Mesh Vertices Uncertainty Regression model (see Section \ref{sec:AMVUR} for more details). 
The dependencies between the variables $\Jb^{i}, \Vb^{i}, \Ja^{i}, C^{i}$ and $I^{i}$ are governed by a Bayesian Network represented via the Directed Acyclic Graph (DAG) shown in Figure \ref{fig:dependence}. Bayes' theorem allows $P\left(\Ja^{i},\Jb^{i},\Vb^{i},C^{i}|I^{i};\delta\right)$ to be factorized using the DAG as the product of $P\left(\Vb^{i}|I^{i};\delta\right)$, 
$P\left(C^{i}|I^{i};\delta\right)$, 
$P\left(\Jb^{i}|\Vb^{i};\delta\right)$ and 
$P\left(\Ja^{i}|\Jb^{i},C^{i};\delta\right)$. 
During training, the probability distribution of $Q\left(\Ja^{i},\Jb^{i},\Vb^{i},C^{i}|I^{i};\theta\right)$ and the prior probability distribution generated by the MANO model are encouraged to be close to each other. This allows the probability distribution training of $Q\left(\Ja^{i},\Jb^{i},\Vb^{i},C^{i}|I^{i};\theta\right)$ to be conditioned on the prior distribution.
$Q\left(\Ja^{i},\Jb^{i},\Vb^{i},C^{i}|I^{i};\theta\right)$ can be factorized as the product of  $Q\left(\Vb^{i}|I^{i};\phi\right)$, 
$Q\left(C^{i}|I^{i};\gamma\right)$, $Q\left(\Jb^{i}|\Vb^{i};\phi\right)$ and $Q\left(\Ja^{i}|\Jb^{i},C^{i};\phi,\gamma\right)$, where $\phi,\gamma \in \theta$ are trainable parameters in our AMVUR model.

Considering observation noise and model error, we assume the approximate probability distribution of $Q\left(\Vb|I;\phi\right)$ and prior probability distribution of $P\left(\Vb|I;\delta\right)$ take on Gaussian distributions $\mathcal{N}_{\phi}\left(\mu_{\phi},diag\left(\sigma_{\phi}\right)\right)$ and $\mathcal{N}\left(\mu_{v},diag\left(\sigma_{v}\right)\right)$, respectively. $\mu_{\phi},\sigma_{\phi}\in \mathbb{R}^{\mathcal{V}\times 3}$ and $\mu_{v},\sigma_{v}\in \mathbb{R}^{\mathcal{V}\times 3}$ are learned from our AMVUR model and the MANO model. Given the above, we can derive our loss function for mesh vertices as:
\begin{equation}
\small
\begin{split}
\mathcal{L}_{\Vb} &
=-\ln Q\left(\Vb^{i}|I^{i};\phi\right) + \ln \frac{Q\left(\Vb ^{i}|I^{i};\phi\right)}{P\left(\Vb^{i}|I^{i};\delta\right)}\\
&=\sum_m\left[\frac{1}{2}\left(\frac{\bVb^{i,m}-\mu^{m}_{\phi}}{\sigma^{m}_{\phi}}\right)^{2}+\ln\sqrt{2\pi}\sigma_{\phi}^{m}\right]\\
&\quad +\frac{1}{2}\left[\ln \prod_{m}\frac{\sigma_{v}^{m}}{\sigma_{\phi}^{m}}-d 
+\sum_{m}\frac{\sigma_{\phi}^{m}+\left(\mu_{v}^{m}-\mu_{\phi}^{m}\right)^{2}}{\sigma_{v}^{m}}\right]
\end{split}
\label{eq:V_3D}
\end{equation}
where $\bVb^{i,m}$ denotes the ground truth 3D coordinates of the mesh vertices of the $i^{th}$ image. $m$ is an index of each dimension. The mean $\mu_{\phi}$ and variance $\sigma_{\phi}$ are learned via two MLP neural networks with $\phi \in \theta$. The mean $\mu_{v}$ of prior net is learned via the MANO model and variance $\sigma_{v}$ is supposed to be equal to $\textbf{1}$. $d$ denotes the dimension of $\mu_{v}$. The last term of the equation penalizes difference between the approximate distribution $Q$ and the prior distribution $P$ during training. \edited{Different from the previously widely used L1/L2 loss, which is less able to capture the data distribution, our loss function allows our model to consider the uncertainty and variability in the hand, which is important for modeling complex and varied 3D meshes. Further, sampling from the distribution during training of our probabilistic model allows the model to explore different variations of the mesh, leading to a more robust and generalizable model.}

Since the prior probability distribution of camera parameters is unknown, we assume that $Q\left(C|\Ja^{i},I^{i};\gamma\right)$ and  $P\left(C|\Ja^{i},I^{i};\delta\right)$ are subject to Gaussian distributions $\mathcal{N}_{\gamma}\left(\mu_{\gamma},diag\left(\textbf{1}\right)\right)$ and $\mathcal{N}\left(\bar{C}^{i},diag\left(\textbf{1}\right)\right)$. The loss function for the camera parameters can then be derived as:
\begin{equation}
\small
\begin{split}
\mathcal{L}_{C}
&=\sum_m\left(\bar{C}^{i,m}-\mu^{m}_{\gamma}\right)^{2},
\end{split}
\label{eq:C}
\end{equation}
where $\bar{C}^{i,m}$ denotes the $m^{th}$ index of the ground truth camera parameters of the $i^{th}$ image. The mean $\mu_{\gamma}$ is learned via a MLP neural network.

To model the dependence between the 3D joints and 3D mesh vertices, we follow the common use in \cite{park2022handoccnet,romero2017embodied, chen2021model, lin2021end} to use a pre-defined regression matrix $B \in \mathbb{R}^{K \times \mathcal{V}}$ from the MANO model. Meanwhile, the loss function for the 3D joints can be derived as:
\begin{equation}
\small
\begin{split}
\mathcal{L}_{\Jb}
&=\sum_{m}\left[\frac{1}{2}\left(\frac{\bJb^{i,m}-\left(B\mu_{\phi}\right)_{m}}{\left(B\sigma_{\phi}\right)_{m}}\right)^2+\ln\sqrt{2\pi }\left(B\sigma_{\phi}\right)_{m} \right]\\
&\quad +\frac{1}{2}\left[\ln\prod_{m}\frac{  \left(B\sigma_{v}\right)_{m}}{\left(B\sigma_{\phi}\right))_{m}}-d\right.\\ &\quad\left.+\sum_{m}\frac{\left(B\sigma_{\phi}\right)_{m}+\left(\left(B\mu_{v}\right)_{m}-\left(B\mu_{\phi}\right)_{m}\right)^{2}}{\left(B\sigma_{v}\right)_{m}}\right],
\end{split}
\label{eq:J_3D}
\end{equation}
where $\bJb^{i,m}$ denotes the ground truth 3D joints.

To model the dependence between the 2D joints, the 3D joints and camera parameters, a weak perspective camera model: $\Ja = s\Jb R+T$ is adopted, where $s$ is the scale, $R \in \mathbb{R}^{3}$ and $T \in \mathbb{R}^{3 }$ denote the camera rotation and translation of camera parameters $C$, respectively. \edited{The camera parameters are in axis–angle representation using radians followed by Rodrigues’ rotation formula to obtain the rotation matrix.}
The loss function for 2D joints is derived as:
\begin{equation}
\small
\begin{split}
\mathcal{L}_{\Ja}
&=\sum_{m}\left(
\frac{1}{2}\left(\frac{\bJa^{i,m}-S_m(\mu_{\phi})}{S_m(\sigma_{\phi})}\right)^{2}
+\ln\sqrt{2\pi} S_m(\sigma_{\phi})\right)\\
&\quad +\frac{1}{2}\left[\ln \prod_{m}\frac{S_m(\sigma_{v})}{S_m(\sigma_{\phi})}-d\right.\\
&\quad\left. +\sum_{m}\frac{S_m(\sigma_{\phi})+\left(S_m(\mu_{v})-S_m(\mu_{\phi})\right)^{2}}{S_m(\sigma_{v})}\right],
\end{split}
\label{eq:J_2D}
\end{equation}
where $S_m(x) = \left(sBxR+T\right)_m$, and $\bJa^{i,m}$ denotes the ground truth 2D joints of the $i^{th}$ image.

\subsection{Weakly Supervised Problem}
\label{sec:weakly-supervised}
With our defined Bayesian model, we are able to study the problem of training our model under the more challenging condition of no 3D ground truth information (such as 3D keypoints, 3D mesh vertices and camera parameters) being available for training. To tackle this problem, we deal with the variables $\Jb$, ${\Vb}$ and $C$ as hidden variables. So we aim to maximise the following with respect to $\theta$:
\begin{equation}
\small
\begin{split}
&\argmax_{\theta}\ln\mathcal{L}\left( \theta \right)  = \argmax_{\theta}\sum_{i}\ln P\left(\Ja^{i}|I^{i};\theta\right)\\&\quad =\argmax_{\theta}\sum_{i}\ln\iiint P\left(\Ja^{i},\Jb,\Vb,C^{i}|\right.\\&\left.\qquad
I^{i};\theta\right)\mathrm{d}\Jb\mathrm{d}\Vb\mathrm{d}C\\
\end{split}
\label{eq:MLE_no_3D}
\end{equation}

However, direct marginalization of eq. (\ref{eq:MLE_no_3D}) is intractable. So, a
variance inference algorithm is developed to compute the penalized maximum likelihood estimation. The total weakly-supervised loss function of our model is:
\begin{equation}
\small
\begin{split}
&\textit{Loss} =E_{\Vb\sim \mathcal{N}_{\phi},C \sim \mathcal{N}_{\gamma}, \Jb \sim \mathcal{N}_{B}}\\&-\ln P\left(\Ja^{i}|\Jb,\Vb,C,I^{i};\delta\right)\\ 
&+D_{KL}\left[Q_{\phi}\left(\Vb|\Ja^{i},I^{i};\phi\right)\parallel P\left(\Vb|\Ja^{i},I^{i};\delta\right)\right]\\
&+D_{KL}\left[Q_{B}\left(\Jb|\Vb,\Ja^{i};\phi\right)\parallel P\left(\Jb|\Vb,\Ja^{i};\delta\right)\right]\\
&+D_{KL}\left[Q_{\gamma}\left(C|\Ja^{i},I^{i};\theta\right)\parallel P\left(C|\Ja^{i},I^{i};\delta\right)\right],
\end{split}
\label{eq:Loss_unsupervised}
\end{equation}
where $D_{KL}\left[Q\parallel P\right]$ denotes the Kullback–Leibler divergence, which measures how the approximate probability distribution of Q is different from the prior probability distribution of P. $E_{\Vb\sim \mathcal{N}_{\phi},C \sim \mathcal{N}_{\gamma}, \Jb \sim \mathcal{N}_{B}} -\ln P\left(\Ja^{i}|\Jb,\Vb,C,I^{i};\theta\right)$ can be computed using eq. (\ref{eq:J_2D}) after sampling $\Vb$, $\Jb$ and $C$ from probability distributions of $\mathcal{N}_{\phi}\left(\mu_{\phi},diag\left(\sigma_{\phi}\right)\right)$, $\mathcal{N}_{\gamma}\left(\mu_{\gamma},diag\left(\sigma_{\gamma}\right)\right)$ and $\mathcal{N}_{B}\left(B\mu_{\phi},diag\left(B\sigma_{\phi}\right)\right)$. The prior probability distributions of $P\left(\Jb|\Vb,\Ja^{i};\delta\right)$ and $P\left(\Vb|\Ja^{i},I^{i};\delta\right)$ are learned via prior-net. We adopt the camera model of \cite{chen2021model} to estimate the prior probability distributions of $P\left(C|\Ja^{i},I^{i};\delta\right)$, which is assumed to follow the Gaussian distributions $\mathcal{N}_{\gamma}\left(\mu_{\gamma},diag\left(\textbf{1}\right)\right)$. The detailed derivation can be found in the Supplementary Information.

\subsection{Attention-based Mesh Vertices Uncertainty Regression} \label{sec:AMVUR}
 \begin{figure}
\begin{center}
\includegraphics[width=8.5cm]{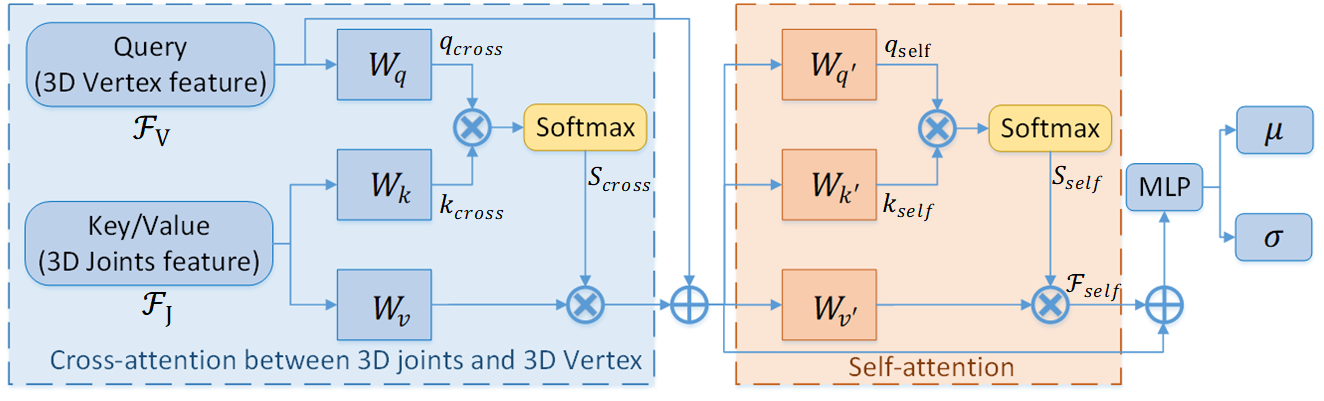}
\end{center}
\vspace{-1.2em}
\caption{The Attention-based Mesh Vertices Uncertainty Regression (AMVUR) module. }
\vspace{-1.2em}
\label{fig:AMVUR}
\end{figure}

The vast majority of previous works \cite{ge20193d,kulon2020weakly,chen2021model,park2022handoccnet} focus on adopting the MANO parametric model and consider regression of pose and shape parameters. However, the pose and shape regression is heavily constrained by the MANO parametric model that was constructed using limited hand exemplars. 
To overcome this limitation, we introduce an Attention-based Mesh Vertices Uncertainty Regression model (AMVUR) to relax the heavy reliance on the MANO model’s parameter space and establish correlations between joints and meshes. To better guide our proposed AMVUR model during training, our probabilistic model takes the MANO parametric model as a prior-net and the AMVUR model estimates the probability distribution of mesh vertices conditioned on the prior-net. The illustration of AMVUR is shown in Figure \ref{fig:AMVUR}.

To construct the 3D vertex and 3D joint features, we firstly extract global feature $\mathcal{F}$ from the backbone CNN. Inspired by the positional encoding of \cite{vaswani2017attention}, we encode positional information by attaching the initial MANO 3D coordinates of joints and mesh vertices to the image-level feature vector $\mathcal{F}$ to obtain the new joint feature matrix $\mathcal{F}_{J}\in \mathbb{R}^{2051 \times K}$ and new vertex feature matrix $\mathcal{F}_{V}\in \mathbb{R}^{2051 \times \mathcal{V}}$. The initial MANO 3D coordinates are obtained by sending zero vectors of pose and shape to the MANO model. Unlike the traditional Transformer method that is only dependant on a self-attention mechanism, we also exploit the correlation between 3D joints and 3D vertices via our cross-attention module. In our cross-attention module, we take 3D vertex features as query and 3D joint features as key and value to model their correlation. With $S_{cross} \in \mathbb{R}^{\mathcal{V}\times K}$ representing the correlation map, we have
\begin{equation}
\small
\begin{split}
S_{cross} = softmax\left(\frac{q_{cross}k_{cross}^T}{\sqrt{d_{cross}}}\right)
\end{split}
\label{eq:cross}
\end{equation}
where $q_{cross} = W_{q}\mathcal{F}_{V}$ and $k_{cross} = W_{k}\mathcal{F}_{J}$ denote the query and key embedding. $d_{cross}$ denotes the feature dimension of the
key $k_{cross}$. The output of the cross-attention module is computed as $\mathcal{F}_{cross}= W_{v}\mathcal{F}_{J}S_{cross}$. $W_{q},W_{k},W_{v}\in \theta$ are trainable parameters applied for query, key and value embedding. We further add a residual connection between $\mathcal{F}_{cross}$ and the primary feature $\mathcal{F}_{V}$, which preserves essential information for mesh vertices regression.

In our self-attention module, we extract the query $q_{self}$, key $k_{self}$ and value $v_{self}$ from $\mathcal{F}_{V}$ by introducing $W_{q'}, W_{k'}, W_{v'}$. We use softmax to generate the correlation map after matrix multiplication of $q_{self}$ and $k_{self}$:
\begin{equation}
\small
\begin{split}
S_{self} = softmax\left(\frac{q_{self}k_{self}^T}{\sqrt{d_{self}}}\right),
\end{split}
\label{eq:self}
\end{equation}
where the self-attention module output is computed as $\mathcal{F}_{self}= W_{v'}\mathcal{F}_{J}S_{cross}$. We add a residual connection between $\mathcal{F}_{self}$ and the output feature $\mathcal{F}_{cross}$ of the previous cross-attention module. Finally two MLP layers are adopted to estimate $\mu$ and $\sigma$ to represent the gaussian probability distribution of 3D coordinates of the mesh vertices.

\subsection{Occlusion-aware Hand Texture Regression}\label{sec:OHTR}
 \begin{figure}
\begin{center}
\includegraphics[width=7cm]{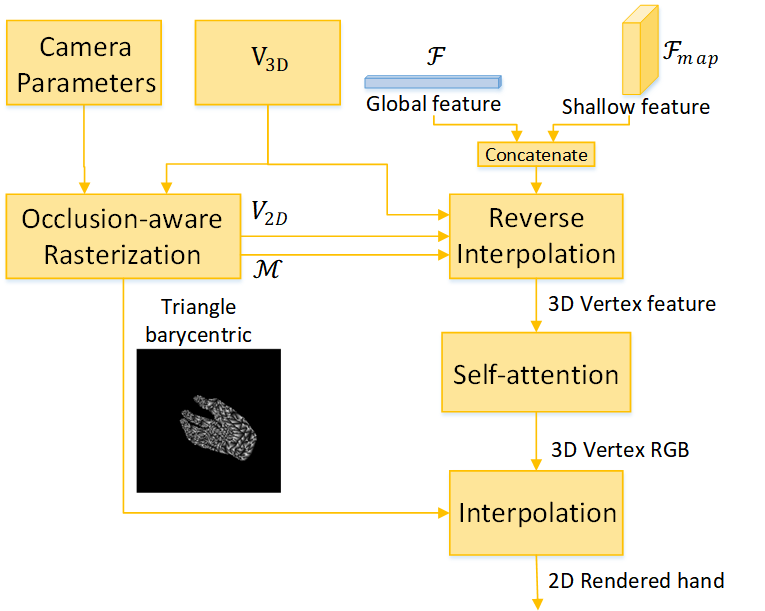}
\end{center}
\vspace{-1.2em}
\caption{The Occlusion-aware Hand Texture Regression module. }
\vspace{-1.2em}
\label{fig:OHTR}
\end{figure}

The hand texture estimation has recently received more attention due to its significant application in immersive virtual reality. However existing hand texture model is unable to generate high-fidelity hand texture and be aware of occlusion. To address above problems, we propose an Occlusion-aware Hand Texture Regression.  As shown in Figure \ref{fig:OHTR}, we regress per-vertex RGB values to represent hand texture. To achieve this goal, we first leverage a rasterizer to implement the mapping between world coordinates and camera pixel coordinates. In our rasterization, a manifold triangle mesh with vertices predicted by our AMVUR model is first created to represent the hand surface. All triangles are then projected to the 2D space, meanwhile per-pixel auxiliary data including barycentric coordinates and triangle IDs are preserved in the rasterization operation. We retrieve visible triangle IDs and create an binary occlusion mask by looking up the three vertices from each visible triangle. Unlike traditional interpolation in Render, which  expands per-vertex data from 3D to pixel space, our reverse interpolation is proposed to construct per-vertex data from pixel to 3D space. The extracted feature on the vertex $\Va^m$ is 
\begin{equation}
\small
\begin{split}
\mathcal{H}_{m}= \mathcal{B}\left(\Va^m,\mathcal{F}\| \mathcal{F}_{map}\|\Vb^m\right),
\end{split}
\label{eq:reverse_inteplation}
\end{equation}
where $\mathcal{B}\left(V, X\right)$ interpolates on the projected 2D point V from tensor X via bilinear interpolation. $\mathcal{F}_{map}$ is a feature map extracted from a shallow layer of the backbone CNN, which preserves rich pixel-level information. $\|$  is the concatenation operation. Afterwards, the 3D Vertex feature $\mathcal{H}_{m}$ is fed into the self-attention layer described in Eq. \ref{eq:self}, followed by a common interpolation for generating the 2D rendered hand image. We adopt the differentiable rasterization and interpolation from \cite{laine2020modular}, allowing our Occlusion-aware Texture Regression model to be trained in an end-to-end manner. Our loss function for training our texture regression model is 
\begin{equation}
\small
\begin{split}
\mathcal{L}_{tex} = \left \| I_{rend}\odot \mathcal{M}-I\odot\mathcal{M}  \right \|_{2},
\end{split}
\label{eq:texture}
\end{equation}
where $I_{rend}$ is the output image of our texture regression model, and $\odot$ denotes elementwise multiplication. $\mathcal{M}$ is a 2D binary occlusion Matrix that indicates the hand region on the texture map, which is obtained from our rasterization.

\begin{table*}
\footnotesize
\begin{center}
\caption{\edited{Hand reconstruction performance compared with SOTA methods on HO3Dv2 after Procrustes alignment.\cite{chen2022mobrecon}$^\star$ develops a synthetic dataset with 1520 poses and 216 viewpoints during training to overcome the long-tailed distribution of hand pose and viewpoint.}
}
\begin{tabular}{p{2.3cm}|p{2.8cm}|p{1.7cm}|p{1.1cm}|p{1.1cm}|p{1.1cm}|p{1.1cm}|p{1.1cm}|p{1.1cm}}
\hline
Training Scheme& Method& Category & $AUC_{J}\uparrow$ & MPJPE$\downarrow$ & $AUC_{V}\uparrow$  & MPVPE$\downarrow$ & $F_{5}\uparrow$  &$F_{15}\uparrow$ \\
\hline
\multirow{15}*{\textit{Supervised}}
& Liu et al.\cite{liu2021semi} & Model-based& 0.803 & 9.9 &0.810&9.5&0.528&0.956\\
&HandOccNet\cite{park2022handoccnet} & Model-based& 0.819 & 9.1 &0.819&8.8&0.564&0.963\\
&I2UV-HandNet\cite{chen2021i2uv} & Model-based& 0.804 & 9.9 &0.799&10.1&0.500&0.943\\
&Hampali et al.\cite{hampali2020honnotate} & Model-based& 0.788 & 10.7 &0.790&10.6&0.506&0.942\\
&Hasson et al.\cite{hasson2019learning} & Model-based& 0.780 & 11.0 &0.777&11.2&0.464&0.939\\
&ArtiBoost\cite{yang2022artiboost} & Model-based& 0.773 & 11.4 &0.782&10.9&0.488&0.944\\
&Pose2Mesh\cite{choi2020pose2mesh} & Model-free& 0.754 & 12.5 &0.749&12.7&0.441&0.909\\
&I2L-MeshNet\cite{moon2020i2l} & Model-free& 0.775 & 11.2 &0.722&13.9&0.409&0.932\\
& METRO \cite{lin2021end} & Model-free& 0.792 & 10.4 & 0.779 & 11.1 & 0.484 & 0.946\\
& Chen et al.\cite{chen2022mobrecon}$^\star$& Model-free&-&9.2&-& 9.4& 0.538& 0.957\\
&Keypoint Trans\cite{hampali2022keypoint} & Model-free& 0.786 & 10.8 &-&-&-&-\\
&\textbf{Ours(prior-net)} & Model-based& 0.783 &10.9 & 0.77&11.5&0.460&0.936\\
 &\textbf{Ours(AMVUR)} & Model-free& 0.814 &9.3 & 0.813&9.4&0.533&0.958\\
&\textbf{Ours(final)} &Probabilistic& \textbf{0.835} & \textbf{8.3} & \textbf{0.836}&\textbf{8.2}&\textbf{0.608}&\textbf{0.965}\\
\hline
\multirow{5}*{\textit{Weakly-Supervised}}& 
$S^{2}HAND$ \cite{chen2021model} & Model-based& 0.765 &-& 0.769 &-&0.44&0.93\\
&\textbf{Ours(prior-net)} & Model-based& 0.752 & 12.4 & 0.760 &12.0&0.417&0.925\\
&\textbf{Ours(AMVUR)} & Model-free& 0.778 &10.8 &0.698 & 15.1 &0.375&0.907\\
&\textbf{Ours(final)}&Probabilistic&\textbf{0.787}&\textbf{10.3}&\textbf{0.784}&\textbf{10.8}&\textbf{0.48}&\textbf{0.949}\\
\hline
\end{tabular}
\label{tab:HO3D}
\end{center}
\vspace{-2.0em}
\end{table*}

\section{Experiments}
We evaluated our method on three widely used datasets for 3D hand reconstruction from a single RGB image: HO3Dv2\cite{hampali2020honnotate}, HO3Dv3\cite{hampali2021ho} and FreiHAND\cite{zimmermann2019freihand}. We present evaluation of the performance of our method in two scenarios: supervised and weakly-supervised training. Our results on the HO3Dv2 and HO3Dv3 datasets were evaluated anonymously using the online evaluation system.\footnote[1]{HO3Dv2:\href{https://codalab.lisn.upsaclay.fr/competitions/4318?}{https://codalab.lisn.upsaclay.fr/competitions/4318?}}\footnote[2]{HO3Dv3:\href{https://codalab.lisn.upsaclay.fr/competitions/4393?}{https://codalab.lisn.upsaclay.fr/competitions/4393?}}. We also present ablation studies to evaluate the importance of each component of the proposed method. 

\subsection{Datasets}
\noindent\textbf{HO3Dv2}\cite{hampali2020honnotate} is a hand-object interaction dataset which includes significant occlusions. The dataset consists of 77,558 images from 68 sequences, which are split into 66,034 images (from 55 sequences) for training and 11,524 images (from 13 sequences) for testing. Each image contain one of 10 persons manipulating one of 10 objects. 

\noindent\textbf{HO3Dv3}\cite{hampali2021ho} is a recently released hand-object interaction dataset with more images and more accurate annotations than HO3Dv2. It contains 103,462 hand-object 3D pose annotated RGB images, which are split into 83,325 training images and 20,137 testing images.


\subsection{Metrics and Implementation Details}
\label{sec:metricsandimplementation}
\noindent\textbf{Evaluation Metrics.}
To quantitatively evaluate our 3D hand reconstruction, we report average Euclidean distance in millimeters (mm) between the estimated 3D joints/mesh and ground truth (MPJPE/MPVPE), and the area under their percentage of correct keypoint (PCK) curves ($AUC_{J}/AUC_{v}$) for the thresholds between $0$mm and $50$mm. For the 3D mesh, we also report F-score of vertices at distance thresholds of $5$mm and $15$mm by $F_{5}$ and $F_{15}$, respectively. 
Following previous work \cite{ge20193d,kulon2020weakly,chen2021model,park2022handoccnet}, we report 3D metrics after procrustes alignment. 


\begin{table*}
\footnotesize
\begin{center}
\caption{\edited{Hand reconstruction performance compared with SOTA methods on HO3Dv3 dataset after Procrustes alignment.} 
}
\begin{tabular}{p{2.3cm}|p{2.8cm}|p{1.7cm}|p{1.1cm}|p{1.1cm}|p{1.1cm}|p{1.1cm}|p{1.1cm}|p{1.1cm}}
\hline
Training Scheme& Method& Category & $AUC_{J}\uparrow$ & MPJPE$\downarrow$ & $AUC_{V}\uparrow$  & MPVPE$\downarrow$ & $F_{5}\uparrow$  &$F_{15}\uparrow$ \\
\hline
\multirow{4}*{\textit{Supervised}}
&ArtiBoost\cite{yang2022artiboost} & Model-based& 0.785 & 10.8 &0.792&10.4&0.507&0.946\\
&Keypoint Trans\cite{hampali2022keypoint} & Model-free& 0.785 & 10.9 &-&-&-&-\\
&\textbf{Ours(prior-net)} & Model-based& 0.780 & 11.3 &0.781 &11.0 &0.471 &0.931 \\
&\textbf{Ours(AMVUR)} & Model-free &0.803  & 9.8 &0.811 & 9.7&0.528 &0.953 \\
&\textbf{Ours} &Probabilistic&\textbf{0.826}& \textbf{8.7} & \textbf{0.834} & \textbf{8.3}& \textbf{0.593}& \textbf{0.964}\\
\hline
\multirow{5}*{\textit{Weakly-Supervised}}& 
$S^{2}HAND$ \cite{chen2021model} & Model-based& 0.769& 11.5 &0.778 &11.1 &0.448 & 0.932\\
 &\textbf{Ours(prior-net)} & Model-based & 0.759 & 12.1 &0.763 & 11.9&0.422 &0.921 \\
&\textbf{Ours(AMVUR)} &Model-free & 0.778 & 10.9 &0.724 & 13.6&0.403&0.904\\
&\textbf{Ours(final)}& Probabilistic & \textbf{0.789} & \textbf{10.5} & \textbf{0.785} & \textbf{10.7} & \textbf{0.475} & \textbf{0.944}\\
\hline
\end{tabular}
\label{tab:HO3D_v3}
\end{center}
\vspace{-1.5em}
\end{table*}

\noindent\textbf{Implementation details.}
All experiments are conducted on two NVidia GeForce RTX 3090 Ti GPUs. We use the Adam optimizer \cite{kingma2014adam} to train the network with batch size of 32. For all supervised experiments, we use ResNet50 \cite{he2016deep} as our backbone CNN, following \cite{park2022handoccnet,liu2021semi,moon2020i2l,chen2021i2uv,choi2020pose2mesh,hampali2022keypoint}. We use EfficientNet-B0 \cite{tan2019efficientnet} as our backbone in the weakly-supervised setting, following \cite{chen2021model}. We extract the shallow $\mathcal{F}_{map}$ and global $\mathcal{F}$ features from the first convolution layer and the last fully connected layer before the classification layer of the backbone model, respectively. The code is available on github: \url{https://github.com/ZhehengJiangLancaster/AMVUR}. 

\subsection{Comparison with SOTA Methods}
%
\edited{We compare our supervised and weakly supervised methods against existing state-of-the-art methods on HO3Dv2 and HO3Dv3 in Tables \ref{tab:HO3D} and \ref{tab:HO3D_v3} respectively after applying Procrustes alignment on their results.} We present results on FreiHAND in the Supplementary Material. We conduct experiments on three settings of our proposed model: \textbf{(1) Ours(prior-net)}, where we individually train the MANO prior model, \textbf{(2) Ours(AMVUR)}, where we individually train the proposed Attention-based Mesh Vertices Uncertainty Regression, and \textbf{(3) Ours(final)}, where the MANO prior-net is jointly trained with AMVUR. As shown in Tables \ref{tab:HO3D} and \ref{tab:HO3D_v3}, our probabilistic method achieves the best results across all metrics for both the supervised and unsupervised scenarios. 
%
In the weakly-supervised setting, our approach not only achieves the best performance compared to the other state-of-the-art weakly-supervised approaches, but also outperforms some of supervised approaches such as \cite{choi2020pose2mesh,moon2020i2l}. It is interesting to see that even though our AMVUR outperforms the other state-of-the-art Model-free approaches\cite{chen2021i2uv,lin2021end,choi2020pose2mesh} in the supervised training scheme, its contribution is lower in the weakly-supervised training scheme due to the increasing solution space of mesh reconstruction. \edited{Evaluation before Procrustes alignment is reported in the Supplementary Material.
}
Figure \ref{fig:mesh_comp} shows that our probabilistic model generates more accurate hand pose and shape than the other state-of-the-art methods on the HO3Dv2 dataset. Despite some hands being severely occluded, our probabilistic model produces better results.

\begin{figure}
\begin{center}
\includegraphics[width=8cm]{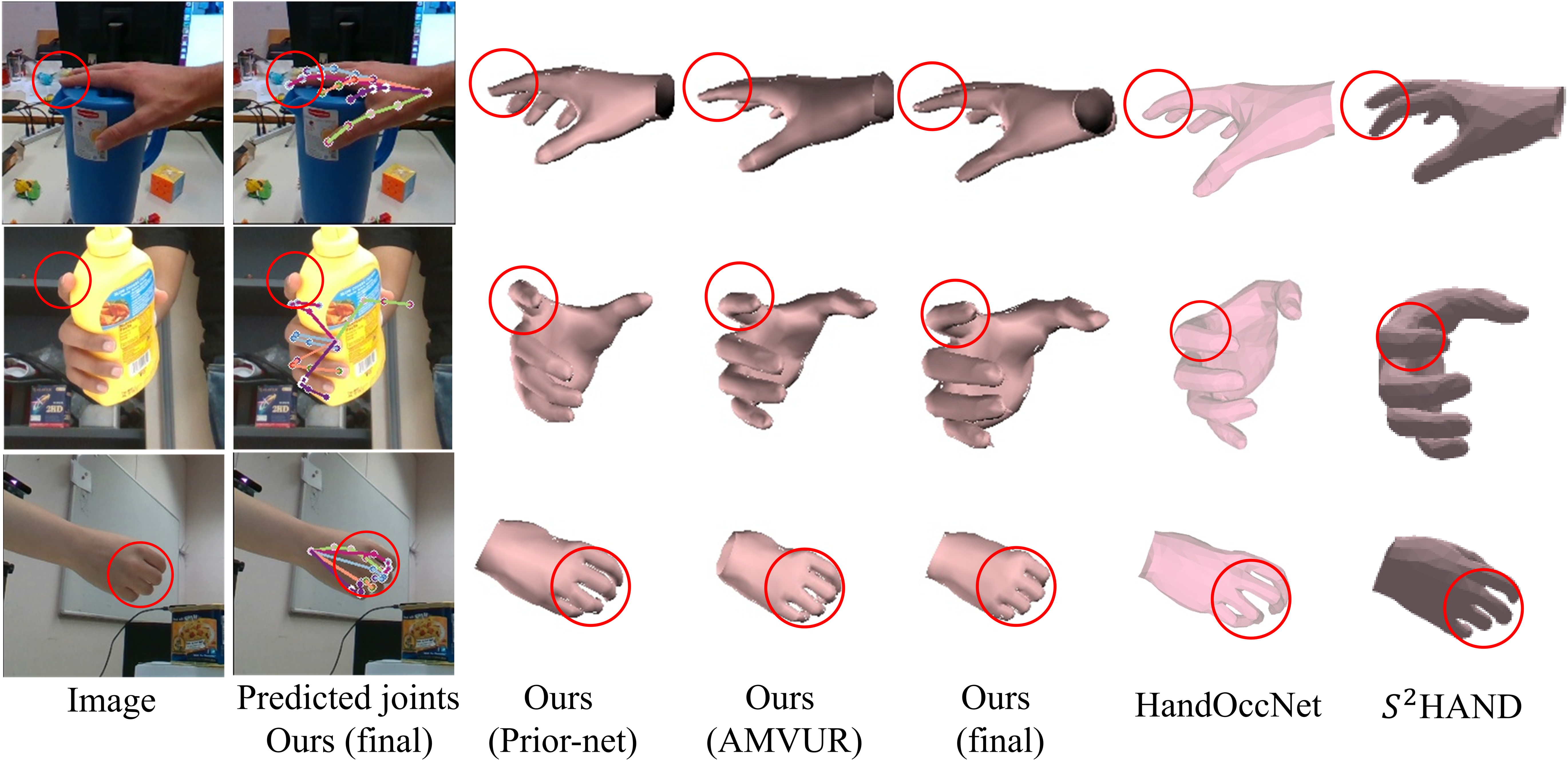}
\end{center}
\vspace{-1.2em}
\caption{Qualitative comparison of the proposed models and SOTA 3D hand mesh estimation methods HandOccNet\cite{park2022handoccnet} and $S^2$HAND \cite{chen2021model} on HO3Dv2.}
\label{fig:mesh_comp}
\end{figure}

\begin{table}
\footnotesize
\begin{center}
\caption{\edited{Impacts of loss terms in the supervised training scheme on HO3Dv2 dataset after Procrustes alignment.}}
\vspace{-0.9em}
\begin{tabular}{p{0.4cm}p{0.4cm}p{0.4cm}p{0.4cm}cccc}
\hline
\multicolumn{4}{c}{Loss terms}& \multirow{2}{*}{MPJPE$\downarrow$} & \multirow{2}{*}{MPVPE$\downarrow$}&\multirow{2}{*}{$F_{5}\uparrow$}&\multirow{2}{*}{$F_{15}\uparrow$} \\
\cline{1-4}
$\mathcal{L}_{\Vb}$&$\mathcal{L}_{\Jb}$&$\mathcal{L}_{C}$&$\mathcal{L}_{\Ja}$& & &&\\
\hline
 \checkmark &  &   & &9.7&8.4&0.596&0.960\\
\hline
 \checkmark & \checkmark &   & &8.5&8.3&0.605&0.964\\
\hline
 \checkmark & \checkmark & \checkmark  &  & 8.6&8.3&0.606&0.963 \\
 \hline
\checkmark & \checkmark & \checkmark  & \checkmark & \textbf{8.3} & \textbf{8.2}&\textbf{0.608}&\textbf{0.965}\\
\hline
\end{tabular}
\label{tab:ablation_sup}
\end{center}
\vspace{-2.5em}
\end{table}

\begin{table}
\footnotesize
\begin{center}
\caption{\edited{Impacts of loss terms in the weakly-supervised training scheme on HO3Dv2 dataset after Procrustes alignment.}}
\vspace{-0.9em}
\begin{tabular}{p{0.4cm}p{0.4cm}p{0.4cm}p{0.4cm}cccc}
\hline
\multicolumn{4}{c}{Loss terms}& \multirow{2}{*}{MPJPE$\downarrow$} &\multirow{2}{*}{MPVPE$\downarrow$}&\multirow{2}{*}{$F_{5}\uparrow$}&\multirow{2}{*}{$F_{15}\uparrow$} \\
\cline{1-4}
$\mathcal{L}_{\Ja}$&$D_{KL}^{C}$&$D_{KL}^{\Vb}$&$D_{KL}^{\Jb}$&& & &\\
\hline
 \checkmark &  &   & &11.9&11.8&0.434 &0.931\\
\hline
 \checkmark & \checkmark &   & &11.6&11.5&0.442&0.935\\
\hline
 \checkmark & \checkmark & \checkmark  & &11.1&11.0&0.46&0.947\\
 \hline
\checkmark & \checkmark & \checkmark&\checkmark &\textbf{10.3}&\textbf{10.8}&\textbf{0.48}&\textbf{0.949}\\
\hline
\end{tabular}
\label{tab:ablation_weak}
\end{center}
\vspace{-2.5em}
\end{table}

\subsection{Ablation Study}
To verify the impact of each proposed component, we conduct extensive ablation experiments on HO3Dv2.
\subsubsection{Effect of Each Loss Term}
In Table \ref{tab:ablation_sup}, we present the ablation study for the supervised training scheme, where the significant contributions of $\mathcal{L}_{V3D}$ and $\mathcal{L}_{J3D}$ to the 3D mesh reconstruction are seen clearly. It is not surprising to see that $\mathcal{L}_C$ and $\mathcal{L}_{J2D}$ do not give a significant contribution to performance improvement in the supervised training scheme, since their purpose is to help learn the 2D projection and rendering. 
In contrast to the supervised training scheme, we only use 2D joints annotation to reconstruct the hand mesh the in weakly-supervised training setting. So our baseline in Table \ref{tab:ablation_weak} only uses $\mathcal{L}_{J2D}$ without any other constraint or prior knowledge. As shown in Table \ref{tab:ablation_weak}, using $D_{KL}^{C}$ , $D_{KL}^{\Vb}$ and $D_{KL}^{\Jb}$ consistently improves all metrics, demonstrating their benefits. Specifically, $D_{KL}^{\Vb}$ and $D_{KL}^{\Jb}$ bring significant improvements to the mesh (MPVPE, $F_{5}$ and $F_{15}$) and joint (MPJPE) reconstruction, respectively.
\subsubsection{\edited{Analysis of AMVUR model.}}
In terms of regressing the 3D vertex coordinates, a naive approach is to regress vertex coordinates with a series of fully connected layers on the top of our CNN backbone. In experiment A of Table \ref{tab:AMVUR_analysis}, we construct our baseline by replacing AMVUR with fully connected layers to estimate the probabilistic distribution of the vertices. From Table \ref{tab:AMVUR_analysis},  AMVUR clearly outperforms this design, demonstrating the importance of capturing the correlation between joints and mesh vertices during regression. Each major component of our AMVUR, i.e., cross-attention, self-attention and positional encoding, is evaluated in experiments B,C and D, respectively. \edited{In B and C, all tokens use different indices to describe their locations following the traditional Transformer.} We observe that cross-attention and self-attention are critical for performance improvement. Positional encoding further improves the performance of our approach.


\begin{table}
\footnotesize
\begin{center}
\caption{Analysis of the AMVUR. 
}
\vspace{-0.1em}
\begin{tabular}{p{0.5cm}|l|p{1cm}|p{1cm}|p{0.6cm}|p{0.6cm}}
\hline
Exp. &Setup & MPJPE$\downarrow$ &  MPVPE$\downarrow$ & $F_{5}\uparrow$  &$F_{15}\uparrow$\\
\hline
A & Baseline &11.4&11.4&0.462&0.932 \\
B & Self-attention &10.5&10.9&0.496&0.948\\
C & B+Cross-attention  &8.9&8.7&0.581&0.958\\
D & C+Positional &  \textbf{8.3} & \textbf{8.2}&\textbf{0.608}&\textbf{0.965}\\
\hline
\end{tabular}
\label{tab:AMVUR_analysis}
\end{center}
\vspace{-1.5em}
\end{table}

\subsubsection{Comparison of Texture Estimation Model}
\begin{figure}
\begin{center}
\includegraphics[width=8.0cm]{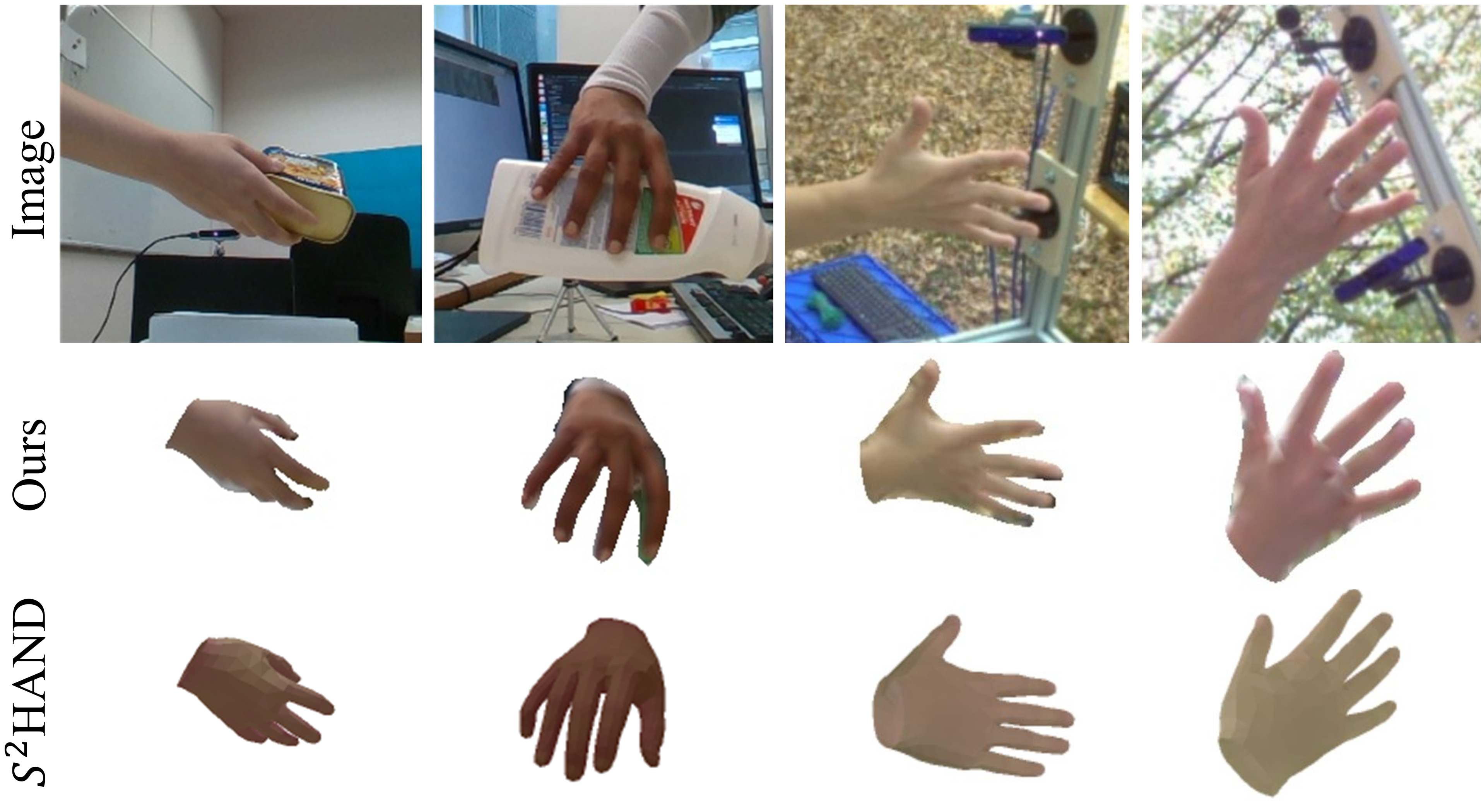}
\end{center}
\vspace{-1em}
\caption{Qualitative comparison of our proposed model and SOTA texture regression model $S^2$HAND \cite{chen2021model} on HO3Dv2.}
\label{fig:texture}
\end{figure}

To quantitatively measure the quality of the estimated hand textures, we use the SSIM \cite{wang2004image} and PSNR \cite{hore2010image} on the hand region. Different from $S^{2}$HAND\cite{chen2021model}, we propose a more intelligent strategy to regress hand texture from the combination of the global and shallow features, which leads to better performance in terms of SSIM and PSNR in Table \ref{tab:texture}. From Fig. \ref{fig:texture}, we can see that the results of $S^{2}$HAND\cite{chen2021model} lack fine details and have larger color differences from the input. In contrast, our approach has better capability in reconstructing high-fidelity hand textures. 
\begin{table}
\begin{center}
\footnotesize
\caption{Quantitative comparison of our model and SOTA texture regression model $S^2$HAND \cite{chen2021model} on HO3Dv2.}
\begin{tabular}{l|p{2cm}|c}
\hline
Method &PSNR$\uparrow$& SSIM$\uparrow$\\
\hline
$S^{2}HAND$\cite{chen2021model}&27.8  &  0.973\\
\textbf{Ours} & \textbf{41.7} & \textbf{0.994}\\
\hline
\end{tabular}
\label{tab:texture}
\end{center}
\vspace{-1.5em}
\end{table}

\section{Conclusion}
In this paper, we have proposed a novel probabilistic model for 3D hand reconstruction from single RGB images, capable of reconstructing not only the 3D joints and mesh vertices but also the texture of the hand accurately despite the presence of severe occlusions. Our approach includes several novelties, including our AMVUR approach, which relaxes the heavy parameter space reliance of the MANO model, allowing more accurate reconstruction. This is trained with our prior-net and includes an attention mechanism to capture correlation between joints and mesh vertices in 3D space. We demonstrated that our proposed probabilistic model achieves state-of-the-art accuracy in fully-supervised and weakly-supervised training. Moreover, we proposed an occlusion-aware Hand Texture Regression model for accurate texture reconstruction.

\noindent
\textbf{Acknowledgments.}
The work is supported by European Research Council under the European Union’s Horizon 2020 research and innovation programme (GA No 787768).

{\small
\bibliographystyle{ieeetr}
\bibliography{cvpr}
}
\end{document}